%% file: main.tex
\documentclass[sigconf]{acmart}

\usepackage{microtype}
\usepackage{graphicx}
\usepackage{subfigure}
\usepackage{multirow} 
\usepackage{booktabs} 
\usepackage[utf8]{inputenc}

\AtBeginDocument{%
  \providecommand\BibTeX{{%
    \normalfont B\kern-0.5em{\scshape i\kern-0.25em b}\kern-0.8em\TeX}}}

\setcopyright{acmlicensed}
\copyrightyear{2024}
\acmYear{2024}
\acmDOI{XXXXXXX.XXXXXXX}

\acmISBN{978-1-4503-XXXX-X/18/06}

\begin{document}

\title{Ada-HGNN: Adaptive Sampling for Scalable Hypergraph Neural Networks}




\author{Shuai Wang}
\affiliation{University of Amsterdam \country{}}

\author{David W. Zhang}
\affiliation{Qualcomm AI Research \country{}}

\author{Jia-Hong Huang \\ Stevan Rudinac \\ Monika Kackovic}
\affiliation{University of Amsterdam \country{}}



\author{Nachoem Wijnberg}
\affiliation{University of Amsterdam \\ College of Business and Economics, University of Johannesburg \country{}}

\author{Marcel Worring}
\affiliation{University of Amsterdam \country{}}



\renewcommand{\shortauthors}{Shuai Wang et al.}

\begin{abstract}
Hypergraphs serve as an effective model for depicting complex connections in various real-world scenarios, from social to biological networks. The development of Hypergraph Neural Networks (HGNNs) has emerged as a valuable method to manage the intricate associations in data, though scalability is a notable challenge due to memory limitations. In this study, we introduce a new adaptive sampling strategy specifically designed for hypergraphs, which tackles their unique complexities in an efficient manner. We also present a Random Hyperedge Augmentation (RHA) technique and an additional Multilayer Perceptron (MLP) module to improve the robustness and generalization capabilities of our approach. Thorough experiments with real-world datasets have proven the effectiveness of our method, markedly reducing computational and memory demands while maintaining performance levels akin to conventional HGNNs and other baseline models. This research paves the way for improving both the scalability and efficacy of HGNNs in extensive applications. We will also make our codebase publicly accessible.

\end{abstract}





\maketitle

\input{parts/01-introduction}

\input{parts/02-related_work}

\input{parts/03-method}
\input{parts/04-experiment}

\input{parts/05-conclusion}

\begin{acks}
We extend our heartfelt thanks to the authors of GRAPES, especially Taraneh Younesian for their in-depth discussions and shared expertise on the deploying of GFlowNet within the realm of graph neural networks. Additionally, we are profoundly grateful to Jiayi Shen, Teng Long for offering their expert insights and analytical perspectives on combinatorial optimization and reinforcement learning.
\end{acks}

\bibliographystyle{ACM-Reference-Format}
\bibliography{sample-base}


\appendix
\section{supplementary materials}

\subsection{Expansion-based propagation on a hypergraph.} An expansion of a hypergraph refers to a transformation that restructures the structure of hyperedges by introducing cliques or pairwise edges, or even new nodes \cite{arya2021adaptive}. This transformation allows traditional graph algorithms to be applied to hypergraphs but has the risk of losing information. Let's take Clique Expansion (CE) as an example. The CE of a hypergraph $\mathcal{G}  \{ \mathcal{V},\mathcal{E} \}$ is a weighted graph with the same set of nodes $\mathcal{V}$ representing $\mathcal{G}$. It can be described in terms of the associated incidence matrice $I^{CE} = I I^T \in \mathbb{R}^{|V| \times |V|}$. One step of node feature $X$ propagation is captured by $I^{(CE)}X$. Then we can formulate message passing process as:

\begin{equation}
    X^{(l+1)}_{v,:} = \sum_{e:v \in e} \sum_{u:u \in e} X^{(l)}_{u,:}
\end{equation}
Many existing hypergraph convolutional layers actually perform CE-based propagation, potentially with further degree normalization and nonlinear hyperedge weights. For example the propagation rule of HGNN~\cite{feng2019hypergraph} takes the following form:
\begin{equation}
    X^{l+1}_{v} = \sigma \left( \left[ \frac{1}{\sqrt{d(v)}} \sum_{e:v \in e} \frac{w_e}{|e|} \sum_{u:u \in e} \frac{X^{(l)}_{u}}{\sqrt{d(u)}} \right] \Theta^{(l)} + b^{(t)} \right), \label{eq:HGNN}
\end{equation}
where $w_e$ means the weight of hyperedge $e$ and can be initialized to be equal for all hyperedges; filter $\Theta$ is the parameter to be learned during the process of extracting features; $\sigma$ is the activation function. The hypergraph convolution implemented in~\citet{HCHA_2021} further leverages weights influenced by node and hyperedge features based on the above paradigm. HyperGCN transforms hyperedges into quasi-clique through the introduction of intermediary nodes, termed mediators~\cite{yadati2019hypergcn}. This method becomes equivalent to a standard weighted Clique Expansion (CE) when dealing with 3-uniform hypergraphs. The hypergraph neural networks we've discussed adapt their propagation using CE-based methods, sometimes with non-linear hyperedge weights, showing notable effectiveness on standard citation and coauthorship benchmarks.

\subsection{GflowNet}
Generative Flow Networks~\cite{bengio2021flow, bengio2023gflownet, madan2023learning, liu2023gflowout, hu2023gflownetem} aims to train generative policies that could sample compositional objects $x \in D$ by discrete action sequences with probability proportional to a given reward function. It is a combination of Reinforcement Learning, Active learning, and Generative Modelling. This network could sample trajectories according to a distribution proportional to the rewards, and this feature becomes particularly important when exploration is important. The approach also differs from RL, which aims to maximize the expected return and only generates a single sequence of actions with the highest reward. GFlowNets has been applied in molecule generation~\cite{bengio2021flow}, discrete probabilistic modeling~\cite{zhang2022generative}, robust scheduling~\cite{zhang2022robust}, Bayesian structure learning ~\cite{deleu2022bayesian}, high-diemensional sampling~\cite{zhang2023diffusion} and causal discovery~\cite{li2023gflowcausal}, as well as graph sampling and generation~\cite{younesian2023grapes, li2023dag}.

\textbf{Trajectory Balance Loss}
\citet{malkin2023trajectory} proposes to estimate $Z$, $P_F$, and $P_B$ by optimizing the trajectory balance loss which is the squared difference between the two parts of the equation \ref{eq:1}. 
\begin{equation}
    \mathcal{L}_{TB}(\tau) = \left( \log \frac{Z(s_0) \prod_{t=1}^{n} P_F(s_t|s_{t-1})}{R(s_n) \prod_{t=1}^{n} P_B(s_{t-1}|s_t) }\right)^2.\label{eq:TBL}
\end{equation}
To apply the trajectory balance loss in the conditional case, we would need to learn an additional regression model that estimates the log-partition function $log Z$ conditioned on $G_l$. Training such a network accurately is difficult but crucial for learning the probability $P_F$. In particular, a wrong estimation of $logZ$ can incorrectly change the direction of the gradients of the loss function. This loss assumed that the normalizing function $Z(s_0)$ in Equation \ref{eq:TBL} is constant across different mini-batches to reduce estimation overhead. However, a wrong estimation of $log Z$ can incorrectly change the direction of the gradients of the loss.  

\subsection{Hypergraph}
Hypergraphs can be used in many cases. For example in scientometrics~\cite{li2024high} to study various aspects of scientific research and its impact. For example, a hypergraph could be used to represent the relationships between different scientific papers, where a vertex represents each paper, and each hyperedge represents a relationship between two or more papers (e.g., co-authorship or co-citation relationships). By analyzing the structure and properties of the hypergraph, it is possible to gain insights into the patterns and trends within the scientific community, such as the level of collaboration among researchers, the impact of different research areas, and the influence of different institutions. Hypergraphs can also be used to visualize the relationships between topics or fields and to identify areas of overlap and potential connections between different areas of study. Besides, hypergraph also has the potential to enhance other applications like Multimedia retrieval or generation~\cite{9222341, HyperLearn, Arya_Relational, zhu2024enhancing, li2021paint4poem, wang2023active}

\end{document}

%% file: parts/01-introduction.tex
\section{Introduction}


Geometric deep learning on graphs, and graph neural networks (GNNs) in particular, have recently garnered significant attention from the research community. Unlike regular graphs, which model pairwise relations between the nodes using edges, hypergraphs as their mathematical generalization utilize hyperedges that connect multiple nodes, thereby encapsulating higher-order relationships. This makes hypergraphs an appropriate and intuitive framework for representing complex relational structures across various domains. In settings such as social networks \cite{pmlr_chitra19a}, product networks \cite{cheng2022ihgnn, hyper_recom}, and biological networks \cite{pcbi5}, hypergraphs excel in capturing the complex relations and high-order information that regular graphs often cannot. 
Recently, the emergence of HyperGraph Neural Networks (HGNNs) has offered a promising approach to address challenges associated with hypergraph data~\cite{ chen_explainable_2022, yang_clustering_2022, wei_dynamic_2022-1, he_click-through_2021, li_hyperbolic_2021,proto_HGNN, yu_supervised_2022}. However, HGNN-based methods face scalability challenges due to the requirement of storing complete incidence matrices and feature matrices in memory, like information for millions of nodes and hyperedges for academic networks. This results in significant memory consumption and extended training time, making the direct application of HGNNs to large hypergraphs impractical~\cite{Antelmi2023}.

\begin{figure}[!t]
	\centering
	\includegraphics[width=\columnwidth]{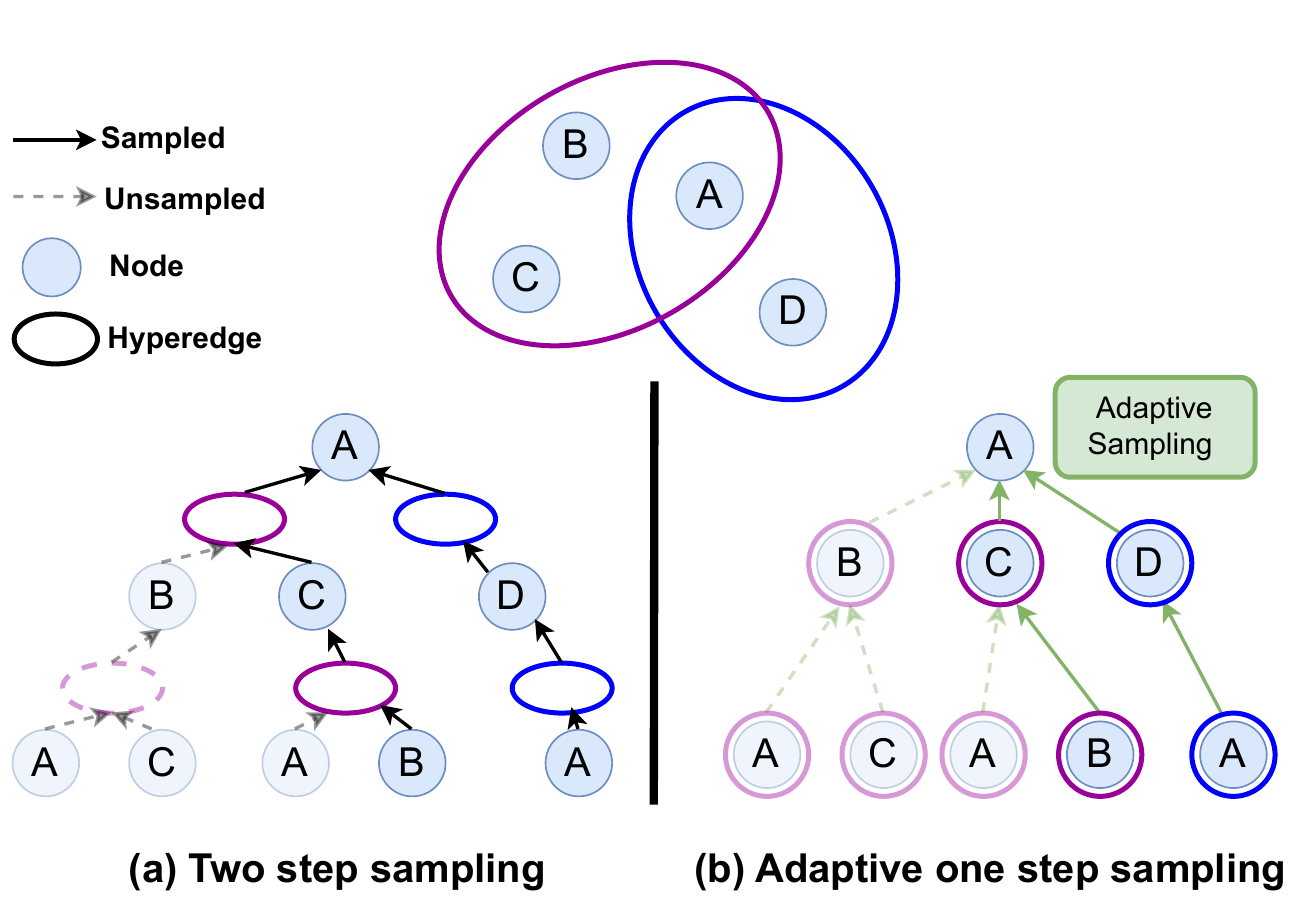}\vspace{-2mm}
	\caption{A visual comparison between the standard 2-hop HGNN computation sampling and our proposed adaptive sampling method.  
        (a) This diagram depicts the conventional HGNN computation, which sequentially processes information from node to hyperedge and back to node, necessitating a two-step sampling strategy that accounts for both nodes and their connecting hyperedges.   
        (b) Illustrates our proposed method, which streamlines this process by integrating nodes with their hyperedges into a singular vertex representation and employing adaptive sampling of neighbor nodes that are relevant to the task at hand. This approach aims to reduce memory overhead while preserving task performance.}
    \label{fig:example}
\end{figure}

Sampling techniques, commonly utilized across various tasks ~\cite{pmlr-v28-meng13a, duan2023comprehensive, hamilton2018inductive, chen2018stochastic, chen2018fastgcn}, have demonstrated effectiveness in mitigating the scalability challenge. By working with a subset of the data rather than the entire dataset, these techniques significantly reduce memory consumption and enable scalability. 
A particular two-step sampling technique, illustrated in Fig.~\ref{fig:example}(a) and originally designed for regular graphs \cite{hamilton2018inductive, chen2018stochastic}, can be adapted to tackle scalability issues in hypergraphs. However, this adaptation often neglects the complex interaction between hyperedges and nodes, as well as the inherent structural complexities unique to hypergraphs. It assumes a degree of similarity in complexity between hypergraphs and regular graphs that may not accurately reflect reality~\cite{yang2020hypergraph, aponte2022hypergraph}. 
Furthermore, employing a sampling technique with a designated sampling policy entails working with only a portion of the data rather than the entirety, leading to inherent information loss. Take the sampling technique incorporated with a fixed sampling policy as an example~\cite{pmlr_chitra19a}. Such fixed sampling mainly relies on node degree or other features, potentially risking information loss by overlooking less central, yet potentially informative nodes. This limitation undermines the ability of existing HGNNs to effectively learn from a subset of hypergraph data, as structural and semantic information may be lost in the process. Given the complexity inherent in hypergraphs, coupled with the limitations of sampling techniques, there is a pressing need for a more effective approach to address the scalability issue within the hypergraph domain.

To overcome the two aforementioned challenges, this study introduces a novel approach that integrates a specially designed one-step adaptive sampling technique. 
This one-step design selects individual nodes while also considering the multi-node connections represented by hyperedges. In addition, as shown in Fig.\ref{fig:example} (b), the proposed adaptive mechanism selectively samples from both hypernodes and hyperedges, effectively integrating node attributes with their corresponding hyperedges. Guided by reward-driven learning, this adaptive mechanism iteratively selects samples to retain essential hypergraph features. Notably, this fused node-hyperedge representation maintains higher-order relationships by establishing interconnectedness within the expanded space. Our proposed one-step adaptive hypergraph sampling method shown in  Fig.~\ref{fig:framework} ensures diversity, facilitating exploration while maintaining performance to preserve the hypergraph's structural and semantic details. Consequently, our method effectively captures both the intricate hyperedge context and the nodes within, ensuring efficient learning for the rich and interconnected context provided by hyperedges.

Moreover, the introduced adaptive mechanism may inadvertently adapt to noise or outliers in the data, rather than capturing the underlying patterns or structure. This can lead to instability and reduced robustness, as the adaptive mechanism may overreact to fluctuations in the data. To enhance the robustness of our proposed adaptive sampling method and counteract overfitting to the existing topology, we introduce a technique called Random Hyperedge Augmentation. This strategy enriches the search space for adaptive sampling, fostering greater generalization across unseen topologies and bolstering the robustness of the sampling process. 
Concurrently, to expedite convergence during large-scale training under the adaptive sampling scheme, we introduce a supplementary Multilayer Perceptron (MLP) module. Leveraging its scalability and fast training characteristics, this MLP module is exclusively trained on the node features of the hypergraph dataset, omitting topological data to ensure swift learning. The trained MLP then acts as a pretraining foundation,  speeding up the subsequent training process of our HGNN models. These methodological enhancements collectively contribute to a more efficient and effective hypergraph learning architecture.

We conduct extensive experiments to evaluate our method across various real-world datasets, showcasing its ability to significantly decrease computational and memory costs while maintaining and even surpassing traditional full-batch HGNNs and other baseline methods in node classification tasks. Through the development of this innovative learning-based approach to hypergraph sampling, this work unveils a new avenue for enhancing the scalability and effectiveness of HGNNs for large-scale applications, thereby broadening their practical utility.


\vspace{+3pt}
\noindent Our primary contributions can be summarized as follows:

\begin{itemize}
   \item We address the Scalability in Hypergraph Learning by considering a computational perspective of message passing and designing the sampling policy accordingly.

   \item We introduce a novel one-step adaptive sampling technique designed specifically for hypergraph learning, which uniquely accommodates its complexities by considering both individual nodes and their multi-node connections via hyperedges.

    \item To enhance the robustness of training, we integrate a Random Hyperedge Augmentation technique to enrich the search space for adaptive sampling. Additionally, a supplementary MLP module pre-trained on node features is proposed to accelerate the training of HGNN models.

    \item To demonstrate the efficacy and effectiveness of our proposed method, we conducted extensive experiments across seven real-world datasets, the largest of which contains over one million hyperedges, 

\end{itemize}

%% file: parts/02-related_work.tex
\section{Related work}
In this section, we will review related work on hypergraph expansion and neural networks, and scaleable hypergraph neural networks.
 \subsection{Hypergraph expansion and neural networks}
 Hypergraphs are often converted into regular graph structures through various expansion methods, such as clique~\cite{CE_08}, star~\cite{Agarwal_spectral}, and line expansions~\cite{yang2023semisupervised}. ~\citet{Agarwal_spectral} provides a comprehensive spectral analysis of different higher-order learning methods utilizing these expansions. Despite the simplicity of expansions, it is well-known that they may cause distortion and lead to undesired losses in learning performance~\cite{pmlr-v89-chien19a,pmlr-v80-li18e}. In the context of predictive modeling, earlier spectral-based hypergraph neural networks are analogous to applying GNNs on clique expansion. This is evident in models such as HGNN~\cite{feng2019hypergraph}, HCHA~\cite{HCHA_2021}, and H-GNNs~\cite{zhang2022hypergraph}. Meanwhile, HyperGCN~\cite{yadati2019hypergcn} approaches the reduction of hyperedges by employing Laplacian operators, representing a variant of the clique expansion methodology. Recently,  researchers have proposed several models such as HyperSAGE~\cite{arya2020hypersage}, UniGNN~\cite{huang2021unignn}, and AllSet~\cite{chien2022you} that employ a vertex-hyperedge-vertex information propagating pattern to iteratively learn data representations. They can be interpreted as GNNs applied to the star expansion graph, where hyperedges are also represented as nodes. However, these models commonly face significant challenges in time and memory complexity when learning on large hypergraphs data, primarily due to the necessity of full batch training.

\subsection{Scalable hypergraph neural networks}
To develop scalable hypergraph learning methods, it is beneficial to examine how this is achieved in graph neural networks. Various approaches have been proposed to enhance the scalability of Graph Neural Networks (GNNs)~\cite{duan2023comprehensive}. 
Previous works in this area can be categorized into two branches: sampling-based and decoupling-based. In general, decoupling-based methods, which require large CPU memory space, are still not feasible for very large graphs~\cite{duan2023comprehensive}. We therefore focus on sampling-based approaches. The sampling algorithms for GCNs broadly fall into three categories: node-wise sampling, layer-wise sampling, and subgraph-wise sampling.  In the ``early'' GNN architectures designed for large graphs, node-wise sampling was predominantly used. This includes methods like GraphSAGE ~\cite{hamilton2018inductive} and Variance Reduction Graph Convolutional Networks (VR-GCN) ~\cite{chen2018stochastic}. 
Subsequently, layer-wise sampling algorithms were introduced to counter the neighborhood expansion issue that arises during node-wise samplings, such as FastGCN ~\cite{chen2018fastgcn}, ASGCN~\cite{huang2018adaptive} and GRAPES~\cite{younesian2023grapes}. In their papers, FastGCN ~\cite{chen2018fastgcn} and ASGCN~\cite{huang2018adaptive} developed a layerwise sampling policy based on importance and variance reduction. GFlowNets, on the other hand, have recently been used for subgraph sampling, integrating an adaptive sampling trajectory to improve the scalability of graph learning and optimization processes ~\cite{zhang2022robust, younesian2023grapes}.
 Graph-wise sampling paradigm is another category, including Cluster-GCN~\cite{Chiang_2019} and GraphSAINT~\cite{zeng2020graphsaint}. Nevertheless, the representation learning for large hypergraphs remains underexplored. An exception is PCL~\cite{kim2023datasets}, a scalable hypergraph learning method that splits a given hypergraph into partitions and trains a neural network via contrastive learning. While it can handle large hypergraph datasets, its learning is dependent on the quality of the partition algorithm based on topology, which inevitably results in information loss across different partitions
To the best of our knowledge, our method is the first scalable hypergraph learning method in which the sampling procedure is adaptive to maintaining and improving task performance.

%% file: parts/03-method.tex
\section{Method}
\begin{figure*}[ht]
    \centering
    
    \includegraphics[width=0.95\linewidth]{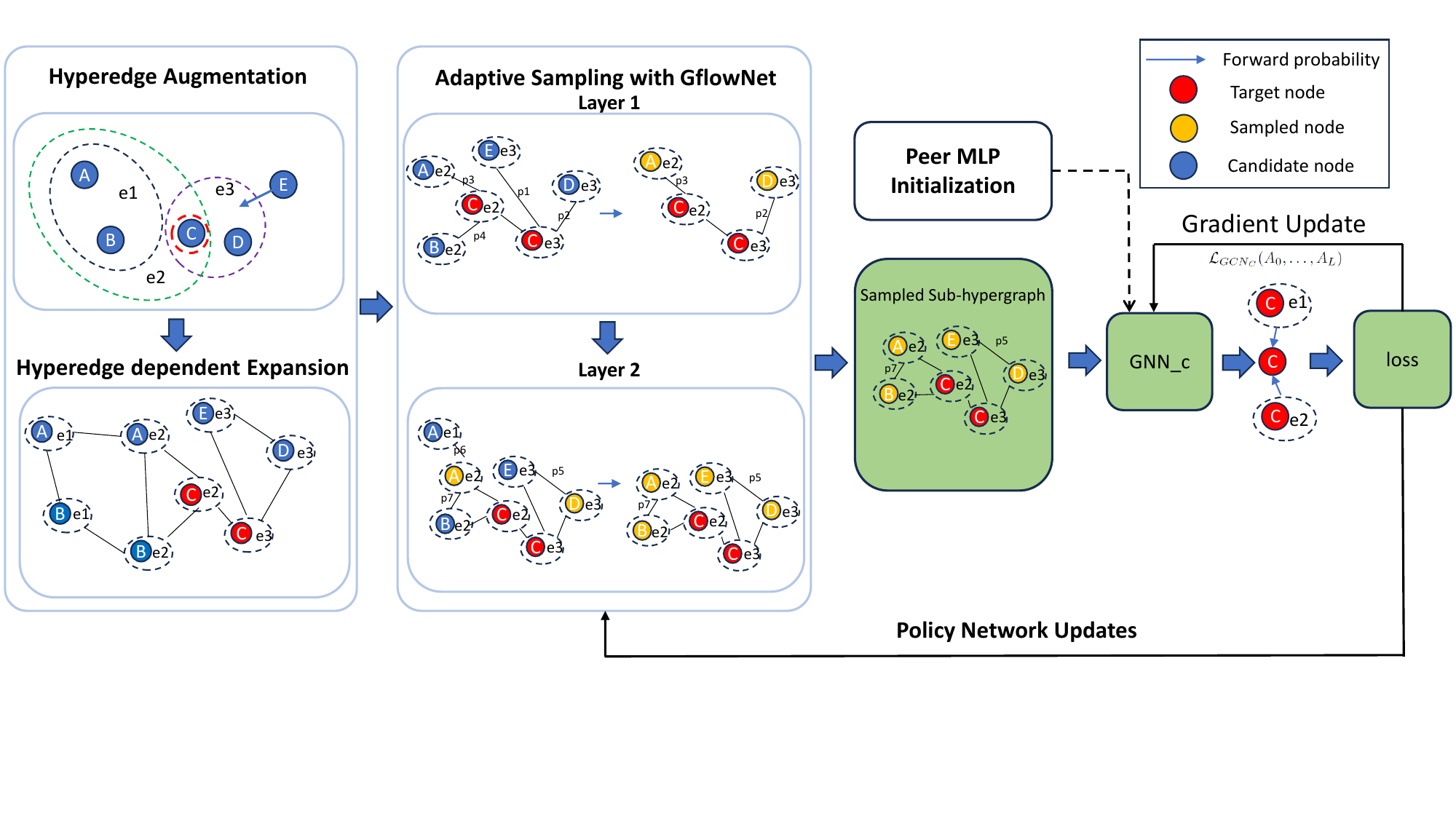}
    \vspace{-20mm}
    \caption{Schematic illustration of the Adaptive Hypergraph Sampling and Learning Process. The workflow begins with a Hyperedge Augmentation and Hyperedge-dependent Expansion, followed by probabilistic node sampling via the GFlowNet policy network. A pre-trained Multi-Layer Perceptron (MLP) can be deployed for initialization of the GNN classifier, which is then processed for gradient update. The GFlowNet is trained using GNN classifier loss as a reward and minimizes log-partition variance, based on trajectory feedback.}
    \label{fig:framework}

\end{figure*}

In this section, we present our method of Adaptive Sampling for Scalable Hypergraph Learning. Specifically, we begin with a brief overview of some critical preliminaries for hypergraph learning in Sec.~\ref{sec:notations}. We further provide a detailed exposition of the proposed adaptive sampling method in Sec.~\ref{sec:expension} and Sec.~\ref{sec:GFlownet}. Then, we discuss the Random hyperedge augmentation and Graph neural networks utilized in Sec.~\ref{sec:RHA} and  Sec.~\ref{sec:GNN}. Fig.~ \ref{fig:framework} presents the overall framework of the proposed Adaptive-HGNN.

\subsection{Hypergraph notations}
\label{sec:notations}
\textbf{Hypergraphs} A hypergraph is represented as $\mathcal{G} = \{ \mathcal{V},\mathcal{E} \}$, where $\mathcal{V} = \{ v_1, v_2,...,v_n \}$ is the node-set, $\mathcal{E}= \{ e_1, e_2,...,e_m\}$ represents the set of hyperedges, and each hyperedge $e\subseteq\mathcal{E}$ consists of 2 or more nodes. A graph is thus a special case of a hypergraph with $|e|$=2, where $|.|$ denotes cardinality of the set, for all hyperedges. This is denoted as a 2-regular hypergraph and more general, $p$-uniform hypergraphs indicate that each hyperedge brings together exactly $p$ vertices.  The relationship between nodes and hyperedges can be represented by an incidence matrix $ I \in  \mathbb{R}^{|\mathcal{V}| \times |\mathcal{E}| }$, with entries defined as:
\begin{equation}
    I(v, e) = 
    \begin{cases}
    1, & \text{if}\ v \in e \\
    0, & \text{otherwise}
    \end{cases}
\end{equation}
For each node $v \in \mathcal{V}$ and hyperedge $e \in \mathcal{E}$, $d(v) = \sum_{e\in \mathcal{E}} I(v,e)$, and $\delta(e) = \sum_{v \in \mathcal{V}} I(v,e)$ denote their respective degree functions. In a hypergraph $\mathcal{G}$, the vertex-degree matrix $D_v \in \mathbb{R}^{|\mathcal{V}| \times |\mathcal{V}|}$ and edge-degree matrix $D_e \in \mathbb{R}^{|\mathcal{E}| \times |\mathcal{E}|}$ are both diagonal, where $D_v$'s diagonal entries $D_{v_{ii}}$ denote vertex degrees and $D_e$'s $D_{e_{ii}}$ denote hyperedge degrees, with non-diagonal entries being zero. The d-dimensional node feature matrix can be defined as $X \in \mathbb{R}^{|\mathcal{V}| \times d}$.

\subsection{Hyperedge-dependent expansion}
\label{sec:expension}
In this section we explain our method step by step as shown in Fig.~\ref{fig:framework}.
The process of information propagation in hypergraph learning can be described as a two-step paradigm. In most related work, the information flows from nodes to hyperedges, and then it is propagated back to the nodes. This pattern is repeated for each propagation step ~\cite{arya2020hypersage, huang2021unignn, chien2022you}. However, this repetitive pattern complicates the process of sampling neighbors in the expanded graph. To address this issue, we propose a different approach for propagation from the view of the computation graph on the target node shown in Fig.~\ref{fig:example}(b). In our approach, we model the information contained within the hypergraph-structured data $\mathcal{G}$ by transforming it into a graph structure where nodes contain both the origin vertex data and the hyperedge information. This transformation simplifies the sampling process by reducing the complexity of decision-making, focusing on individual nodes. Unlike other transformations~\cite{CE_08,Agarwal_spectral}, this approach is reversible and retains the high-order information from the original hypergraph

The graph induced by the expansion of a hypergraph is denoted as $ \mathcal{G}_l = ( \mathcal{V}_l,  \mathcal{E}_l)$. In this graph, the node set $ \mathcal{V}_l$ is defined by pairs of vertices and hyperedges ${(v,e), v \in  \mathcal{V}, e \in  \mathcal{E} }$ from the original hypergraph. The edge set $ \mathcal{E}_l$ and adjacency matrix $ \mathcal{A}_l \in \{0,1\}^{|V_l| \times |V_l|}$ are defined based on the pairwise relation between these pairs. Specifically, $A_l(u_l, v_l) =1$ if either $v = v'$ or $e=e'$ for $u_l = (v ,e )$ and $v_l = (v', e') \in V_l$. With this intuition, we can fuse the hypergraph information by defining the vertex projection matrix without the loss of high-order information. This matrix maps a node $v\in \mathcal{V}$ from the original hypergraph $\mathcal{G}$ to a set of graph vertices $\{v_l=(v, e):e\in \mathcal{E}\}\subset \mathcal{V}_l$ in the induced graph $\mathcal{G}_l$. We introduce the vertex projection matrix $P_{vertex}\in \{0,1\}^{|\mathcal{V}_l|\times |\mathcal{V}|}$,
\begin{equation}
    P_{vertex}(v_l, v) = \left\{
    \begin{aligned}
    ~1&~  & \mbox{if the vertex part of $v_l$ is $v$},\\
    ~ 0&~  & \mbox{otherwise},
    \end{aligned}
    \right.
\end{equation}
Then, given initial feature matrix $X \in \mathbb{R}^{|\mathcal{|V|} \times d_0}$, where $d_0$ is the input node embedding dimension from $\mathcal{G} = (\mathcal{V}, \mathcal{E})$, we transform it into the features in $\mathcal{G}_l = ( \mathcal{V}_l, \mathcal{E}_l )$ by vertex projector $P_{\text{vertex}}$,
\begin{equation}
    H^{(0)} = P_{\text{vertex}}X,  
\end{equation}
$H^{(0)} \in \mathbb{R}^{|\mathcal{V}_l \times d_0|}$ is the initial node feature of the induced graph. This projection transforms vertex features of $\mathcal{G}$ into graph nodes in $\mathcal{G}_l$. Then, we can utilize the technique of neighborhood feature aggregation through graph convolution. This involves incorporating information from both neighboring nodes and hyperedges. The graph convolution for layer $l+1$ can be defined as follows:
\begin{equation}
    h^{(l+1)}_{(v,e)} = \sigma \left(  \sum_{e'} w_{e} h^{(l)}_{(v,e')} \Theta^{(l)} + \sum_{v'} w_{l} h^{(l)}_{(v',e)} \Theta^{(l)} \right),
\end{equation}
where $h^{(l)}_{(v,e)}$ represents the feature representation of the line node $(v,e)$ in the $l$-th layer, $\sigma$ is a non-linear activation function. $\Theta^{(l)}$ is the vector of transformation parameters for layer $l$. Additionally, two hyper-parameters $w_{e}$ and $w_{v}$ are employed to balance the weight from the same node and edge perspective. In summary, our approach redefines the traditional two-step information propagation paradigm by introducing a novel hyperedge-dependent node expansion technique from the perspective of computation graphs. This method simplifies the sampling process in the expanded graph, enabling more straightforward neighbor selection, while retaining the rich, high-order information inherent to the original hypergraph structure.

\subsection{Adaptive sampling with GFlowNet}
\label{sec:GFlownet}
In this section, we describe the adaptive sampling on the expansion hypergraph $\mathcal{G}_l$ with GFlowNet. Before going to applying GFlowNet to hypergraphs, we consider how it is applied in full batch training of GCN~\cite{kipf2017semisupervised}. Given adjacency matrix $A \in \{0,1\}^{|V|  \times |V|}$, the output of the $l^{th}$ layer of GCN can be represented as $H^l = \sigma (\hat{A}H^{l-1}W^l)$. Here, $W^l \in \mathbb{R}^{d^l \times d^{l+1}}$ is the weight matrix of layer $l$, $\hat{A} = \widetilde{D}^{-1}(A+I)$, and $\sigma$ denotes a non-linear activation function. Furthermore, $d^l$ refers to the hidden dimension of layer $l$, where $d^0$ represents the input feature size. For a node $v_i\in \mathcal{V}$, the corresponding update can be described as follows:
\begin{equation}
    h^{l}_{v_i} = \sigma \left( \sum_{v_j \in \mathcal{N}(v_i)} \hat{A}_{(v_i, v_j)} h^{l-1}_{v_j} W^{l} \right),
\end{equation}
where $\mathcal{N}(v_i)$ represents the set of neighbors of $v_i$.

As the number of layers increases, the computation graph for node  $v_i$ becomes more complex, involving neighbors from increasingly distant hops. Additionally, nodes with a larger number of neighbors will further exacerbate this situation, leading to higher computation requirements. In the following paragraph, we will introduce the concept of sampling-based information message passing. To begin, the target nodes are divided into mini-batches of size $b$. Then, in each layer, $k$ nodes are selected from the neighbors of the nodes in the previous layer with certain probabilities. Therefore, the approximation of the $l$-th update of node $v_i$ is as follows:
\begin{align}
    \tilde{h}^l_{v_i} = \sigma \left ( \sum_{v_j \in K^l} \hat{A}'^l_{(v_i, v_j)}\tilde{h}^{l-1}_{v_j}W^l  \right ), \\ \notag
    K^l \sim q(K^l|\mathcal{N}(K^{l-1},...,K^0)) 
\end{align}
where set $K^l \in \mathcal{N} (K^{l-1},...,K^0)$ represents the sampled nodes in layer $l$. Here, $K^0$ refers to the set of mini-batch target nodes. $\hat{A}'^l_{(v_i,v_j)}$ represents the row normalized value of the sampled adjacency matrix $A^l$ in layer $l$. Our objective is to learn the $q(K^l|\mathcal{N}(K^{l-1},...,K^0)$ probability of sampling the set of nodes $K^l$ given the nodes that were sampled.

Generative Flow Networks, also known as GFlownet, is a framework family introduced by~\citet{bengio2021flow} that focuses on training generative policies. These policies can sample compositional objects represented by the variable $x$, which belongs to the set $D$. The sampling is done using discrete action sequences, and the probability of selecting a particular sequence is determined based on a provided reward function. One of the key advantages of GFlowNet is its ability to generate diverse samples. This diversity is crucial as it facilitates exploration and helps mitigate issues such as overfitting. In this work, we use the exponential negative value of classification loss as the reward.

 Let $\mathcal{G}_F = (\mathcal{S}, \mathcal{A}, \mathcal{S}_0, \mathcal{S}_f, \mathcal{R})$ represent a GFlowNet learning problem. In the case of adaptive sampling, trajectories initialize with the root batch of node $\mathcal{S}_0$. We denote the set of all such trajectories as $\mathcal{T}$, and the set of trajectories that end at $x$ as $\mathcal{T}_x$. Additionally, we introduce a flow function $F$:  $\mathcal{T} \rightarrow \mathcal{R}^{+}$ and its associated normalized probability distribution $P(s) = F(s)/Z$, where $Z = \sum_x R(x) $. For any Markovian flow, we can break down the probability of a trajectory in terms of the forward probability: 
\begin{equation}
    P(s) = \prod_{t=1}^{n} P_F(s_t | s_{t-1}). 
\end{equation}

To generate trajectories $s$, we can sample a sequence of actions starting from the initial state $s_0$. Similarly, we can define a backward probability $P_B$ that incorporates the probability of the trajectory:
\begin{equation}
    P(s|s_n = x) = \prod_{t=1}^{n} P_B(s_{t-1} | s_{t}).
\end{equation}
The training objectives discussed in previous studies aim to establish a consistent flow, whereby consistency refers to the requirement that the forward direction matches the backward direction. In order to achieve a consistent flow $F(s)$ for trajectories $s\in \mathcal{T}_x$, it can be expressed in relation to $P_F$ and $P_B$, and must satisfy the following equality:
\begin{equation}
    Z \prod_{t=1}^{n} P_F(s_t|s_{t-1}) = R(x) \prod_{t=1}^{n} P_B(s_{t-1}|s_t) \label{eq:1}
\end{equation}
Based on this equation, \citet{zhang2022robust} proposes to rewrite Equation~\ref{eq:1} implicitly to estimate $log Z$ based on the forward and backward flows of a single trajectory $s$, where $P_F$ and $P_B$ are neural networks with parameters $\theta$:
\begin{align}
    \zeta(s; \theta) = \log R(x) + \sum_{t=1}^{n} \log P_{B}(s_{t-1}|s_t; \theta)  \notag \\
     - \sum_{t=1}^{n} \log P_{F}(s_{t}|s_{t-1}; \theta), \label{eq:2}
\end{align}
The reward function is defined as $R(s_L) = exp( - \mathcal{L}_{gnn_c}/\tau)$ in this context. Our objective is to train the $\mathcal{G}_F$ model to estimate the forward probabilities $P_F(s_{l-1}|s_{l})$, which represent the likelihood of selecting an adjacency matrix $A_{l}$ based on the previous adjacency matrix $A_{l-1}$ for layer $l-1$. The GFlowNet is designed to predict the probability $P_i$ for each node $v_i$ within the neighborhood $K^l$. This probability indicates whether it will be included in $V^{l}$. Based on Eq.~\ref{eq:2}, it is necessary to define $P_B$. However, the state representation $s=(A_0,...,A_l)$ already represents the trajectory taken through $\mathcal{G}_F$ to get to $s$. There is also include identifier for mini-batch sampled nodes as part of the representation~\cite{younesian2023grapes}. Hence, the $\mathcal{G}$ is a tree structure and $P_{B}(s_{t-1}|s_t) =1 $. 
In the ideal case, the function $\zeta(s; \theta)$ should match the constant partition function $Z$ for all trajectories in a computation graph $\mathcal{G}_c$. This is achieved through a loss function that minimizes the squared deviation of $\zeta(s; \theta)$ from its expected value, thereby refining the model's optimization by focusing on variance reduction.
\begin{equation}
    \mathcal{L}_V (s; \theta) = \left( \zeta(s; \theta) - \mathbb{E}_s [\zeta(s; \theta)] \right)^2
\end{equation}
In practice, we use the training distribution to estimate $\mathbb{E}_s [\zeta(s; \theta)]$ with a mini-batch of sampled trajectories. After passing with the GNN classifier and GFlowNet model, the expansion graph will be transferred back by fusing the higher-order information. We define the vertex back-projection matrix $  P'_{\text{vertex}} \in \mathbb{R}^{|V'| \times |V'|} $,

\begin{equation}
    P'_{\text{vertex}}(v, v_l) = 
        \begin{cases} 
        \frac{\frac{1}{\delta(e)}}{\sum_{(v,e') \in \mathcal{V}_l} \frac{1}{\delta(e')}} & \text{if } v = \text{vertex}(v_l), \\
        0 & \text{otherwise.}
        \end{cases}
\end{equation}
In this way, we can uniquely recover the original hypergraph by
\begin{equation}
    Y = P'_{\text{vertex}} H^{(K)} \in \mathbb{R}^{|V| \times d_o},
\end{equation}
where $d_o$ is the dimension of output representation. The complexity of this 1-layer convolution is of $\mathcal{O}(|\mathcal{E}_l|d_i d_o)$ and the convolution operation could be efficiently implemented as the product of a sparse matrix with a dense matrix. Then, the loss is calculated based on the task.

\subsection{Random hyperedge augmentation}
\label{sec:RHA}
To overcome the limitations inherent in existing hypergraph data, which may not capture all potential connections and risk of overfitting, we have developed a novel random hyperedge augmentation method. This method is particularly useful in real-world datasets, such as those derived from co-authorship, where important but unobserved relationships may exist. Our method seeks to bridge this gap by simulating potential unobserved relationships through the random addition of nodes to existing edges. This approach not only yields a hypergraph that mirrors the complexity of real-world networks more closely but also alleviates the learning model's dependency on the completeness of the observed data. The augmentations process can be defined as a random function $\mathcal{A} : \mathcal{V} \times \mathcal{E} \rightarrow \mathcal{E}$ randomly:
\begin{equation}
    \mathcal{A}(v, e) =  e \cup \{v\} \ \text{randomly }
\end{equation}

This process is particularly synergistic with our GFlowNet-based sampling approach, as it allows the model to explore a richer and more varied feature space that could be absent in the original dataset. Random hyperedge augmentation, when paired with the adaptive sampling capabilities of GFlowNets, ensures that the diversity of sampled subgraphs is representative of both observed and potential hypergraph structures. This dual strategy is critical for learning models that need to generalize well to new data, where the task may involve inferring connections not present in the training set, thereby providing a more complete representation of the underlying network

\subsection{Graph neural networks utilized}
\label{sec:GNN}
In our architecture, we integrate two Graph Neural Networks (GNNs): one functions as the classifier and the other as the policy network for GFlowNet. Specifically, we utilize a pair of Graph Convolutional Networks (GCNs) \cite{kipf2017semisupervised} and a pair of Graph Transformer models \cite{shi2020masked}, which represent the predominant methodologies in GNN modeling.

\textbf{Peer MLP initialization}:
Training Hypergraph Neural Networks (HGNNs) on large-scale graphs presents a notable challenge due to the computationally demanding sparse matrix multiplications involved. Alternative approaches like Multi-Layer Perceptrons (MLPs), which bypass these complex calculations in favor of direct node feature utilization, tend to be faster and simpler but often fall short in predictive performance when applied to hypergraph data. Drawing inspiration from the recent MLPINIT model~\cite{han2023mlpinit} in GNN training, we introduce an innovative initialization strategy for HGNNs. Our method harnesses the efficiency of MLPs to accelerate the training process by starting with MLP-derived pre-trained weights.

\begin{equation}
    \text{MLP: } H^{l} = \sigma(H^{l-1}W^{l}_{mlp}),
\end{equation}
we use the weights trained by MLP as initialization for the hypergraph neural network with hyperedge-dependent node expansion.
\begin{equation}
    H^{l} = \sigma(\tilde{A}H^{l-1}W^{l}_{HGNN}),
\end{equation}
where $W^{l}_{HGNN}$ and $W^l_{MLP}$ are trainable weights of $l$-th layer of HGNN and MLP, respectively. In this way, we can get good pre-trained weights for graph-based models on hypergraph expansion.

\subsection{Training}
We now introduce the training objective for node classification. In node classification, the goal is to derive a labeling function, denoted as $ f : \mathcal{V} \rightarrow {1, 2, \ldots, C} $ based on both labeled data and the geometric structure of the hypergraph. The ultimate objective is to assign class labels to unlabeled vertices through transductive inference. To achieve this, in this paper we propose a strategy that involves minimizing the empirical risk for a given hypergraph $ G = (\mathcal{V}, \mathcal{E}) $ with a set of labeled vertices $ \mathcal{T} \subseteq \mathcal{V} $ and their corresponding labels.
\begin{equation}
f^* = \arg\min_{f(\cdot|\theta)} \frac{1}{|\mathcal{T}|} \sum_{v_t \in \mathcal{T}} \mathcal{L}(f(v_t | \theta), L(v_t)),
\end{equation}
where \( L(v_t) \) is the ground truth label for node \( v_t \) and cross-entropy error is commonly applied in \( \mathcal{L}(\cdot) \). This work focuses on node classification problems on hypergraphs, but it can be easily extended to other hypergraph-related applications, such as hyperedge prediction in which nodes are brought together into new groupings.


%% file: parts/04-experiment.tex
\begin{table}[!h]
	\centering
    \caption{Statistics of chosen real-world hypergraph datasets}	
     \resizebox{\columnwidth}{!}{\begin{tabular}{l|lllll}
		\toprule
		\textbf{Dataset}
		& \textbf{Nodes}  & \textbf{Hyperedges}  &\textbf{Features} &\textbf{$\mathbf{\overline{|e|}}$}  & \textbf{Class}  \\
		\midrule
        NTU2012       & 2,012     & 2,012     & 2048  & 5   & 67            \\ 
        Mushroom      & 8,124     & 112       & 112   & 136.3   & 2             \\
        ModelNet40    & 12,311    & 12,321    & 2048  & 5   & 40            \\
		20News        & 16,242    & 100       & 100  & 654.5    & 4             \\
        DBLP          & 41,302    & 22,363    & 1,425 & 4.5  & 6             \\   
        Trivago       &172,738   & 233,202    & 300  &  3   & 160           \\
        OGBN-MAG      & 736,389   & 1,134,649 & 128  &  6.3   & 349           \\
		\bottomrule
	\end{tabular}}
	\label{tb:hypergraphstat1}
\end{table}

\section{Experimental Setup}

In this section we provide description of the datasets and baseline approaches used in experimental evaluation, alongside the necessary implementation details.

\begin{table*}[hbt!] 
\setlength\tabcolsep{7.5pt}
	\centering
	\caption{Hypernode classification accuracy on real-world hypergraphs (\%). Rdm and Ada represent random and adaptive sampling. We report the mean and standard deviation over 5 runs. Boldfaced letters are used to indicate the best mean accuracy, and underlining is used for the second. OOM indicates out-of-memory, and N/A designates a model numerical error.}
	\begin{tabular}{ l|lllllll} 
		\toprule
		\textbf{Model} & \textbf{20News}  & \textbf{Mushroom} & \textbf{ModelNet40} &  \textbf{NTU2012} & \textbf{DBLP} & \textbf{TriVago}  & \textbf{OGBN\_MAG} \\
		\midrule
		$\mbox{Clique}_{GCN}$    & OOM                          & 91.25 $\pm$ 0.12      & 85.58 $\pm$ 0.12     & 71.72 $\pm$ 4.44        &  86.72 $\pm$ 0.07            & 13.92 $\pm$ 3.37  & OOM \\
        $\mbox{Clique}_{GAT}$     & OOM                          & 83.51 $\pm$ 0.85      & 83.51 $\pm$ 0.85     & 63.76 $\pm$ 2.11        & 86.84 $\pm$ 0.28             & 14.82$\pm$ 1.92   & OOM \\
		HNHN                      & 78.31 $\pm$ 0.39             & 93.24 $\pm$ 0.69      & 91.06 $\pm$ 0.69     & 71.94 $\pm$ 2.71        & 90.29 $\pm$ 0.05    & OOM  & OOM \\
		HGNN                      & 78.31 $\pm$ 0.55             & 93.59 $\pm$ 0.40      & 90.99 $\pm$ 0.03     & 73.57 $\pm$ 0.50        &90.26 $\pm$ 0.24  &26.58 $\pm$ 0.34  & OOM \\
		HyperGCN                  & 78.91 $\pm$ 0.67 & 47.59 $\pm$ 0.21      & 77.22 $\pm$ 1.12     & 49.83 $\pm$ 1.6         & 77.67 $\pm$ 1.44             &8.92 $\pm$ 0.40   &N/A  \\
        SetTransformer            & 79.34 $\pm$ 4.22             & 99.25 $\pm$ 0.50      & \underline{96.54 $\pm$ 0.23}     & 80.32 $\pm$ 2.05        & 91.14 $\pm$ 0.19             & OOM              & OOM \\
        ED-HNN                    & 78.05 $\pm$ 0.51             & 99.44 $\pm$ 0.30      & 95.91 $\pm$ 0.33     & 77.95 $\pm$ 2.64        & $\mathbf{91.57 \pm 0.20}$             & OOM              & OOM \\
        PhenomNN                  & \underline{79.65 $\pm$ 0.64}             & 99.51 $\pm$ 0.32      & 96.44 $\pm$ 0.11     & 82.44 $\pm$ 0.85        & \underline{91.56 $\pm$ 0.26}             & OOM              & OOM \\
        \midrule 
        Rdm-GCN                   & 77.16 $\pm$ 0.98	& 99.63 $\pm$ 0.13	& 86.99 $\pm$ 1.29	& 72.66 $\pm$ 1.40	& 84.37 $\pm$ 1.30	& 31.23$\pm$ 0.32 &27.64 $\pm$ 0.42 \\
        Rdm-GT                    & 77.30 $\pm$ 1.04     &  99.99 $\pm$ 0.01         &   91.28 $\pm$ 1.60        & 74.35 $\pm$ 0.77         & 86.44 $\pm$ 0.28  &33.41$\pm$ 0.21 &27.81 $\pm$ 0.24 \\
        \midrule
        Ada-GCN                   &78.59 $\pm$ 0.61	             &99.73 $\pm$ 0.06	   & 92.53 $\pm$ 0.62	& \underline{83.53 $\pm$ 0.89}  &	84.76 $\pm$ 0.68 &	34.65 $\pm$ 0.66 & 28.32 $\pm$ 0.32\\
        Ada-GT                    & 78.90 $\pm$ 0.99     &  $\mathbf{100 \pm 0.00}$   & 93.44 $\pm$ 0.67 &   82.38 $\pm$ 0.75        & 88.49 $\pm$ 0.12    &  \underline{37.31 $\pm$ 0.57} & \underline{28.84 $\pm$ 0.34} \\
        Ada-GT+RHA                & $\mathbf{79.78 \pm 0.72}$        &  $\mathbf{100 \pm 0.00}$   & $\mathbf{97.20 \pm 0.39}$ &   $\mathbf{84.42 \pm 1.39}$         & 86.40 $\pm$ 0.35    &  $\mathbf{39.42 \pm 0.63}$ & $\mathbf{30.85 \pm 0.44}$ \\
        \bottomrule
	\end{tabular}
	\label{tab:hypergraphResult}
\end{table*}


\subsection{Datasets} 

The existing hypergraph benchmarks predominantly focus on hypernode classification. In this study, we employ four large publicly available benchmark datasets: NTU2012, Mushroom, ModelNet40, and 20NewsW100 as in \cite{yang2023semisupervised}. In addition, we incorporate three larger datasets, specifically DBLP, Trivago, and OGBN-MAG, from \cite{kim2023datasets}. We randomly distribute the data into training, validation, and test sets, adhering to a split ratio of $40\%$, $10\%$, and $50\%$, respectively, to increase the challenge

\subsection{Baselines} In order to evaluate the performance of our methods, we compare them with several baseline models. These baseline models include:

\begin{itemize}
    \item \textbf{Expansion-based baselines:} Clique Expension with GCN \cite{kipf2017semisupervised} and Clique Expension with GAT~\cite{veličković2018graph}. 

    \item \textbf{Spectral hypergraph neural network-based approaches:} HGNN~\cite{feng2019hypergraph},  HyperGCN ~\cite{yadati2019hypergcn} and  HNHN~\cite{dong2020hnhn}. 

    \item \textbf{Spatial Hypergraph neural network based:} SetTransfromer~\cite{chien2022you}, ED-HNN~\cite{wang2023equivariant} and PhenomNN~\cite{wang2023hypergraph}.

    \item \textbf{Random sampling-based:} We also choose GCN~\cite{kipf2017semisupervised} and GraphTransformer~\cite{shi2020masked} with random sampling on hypergraph expansion denoted as Rdm GCN and Rdm GT. 

\end{itemize}
\textbf{Implementations} 
For our experiments, we configured the following hyperparameters: the model consists of two message-passing layers, each with a dimension of 512, and it was trained over 50 epochs. We used the Adam optimizer with a learning rate of $1 \times 10^{-3}$ and a weight decay of $0$. To ensure robustness and reproducibility, all experiments were conducted five times using multiple random splits to calculate the mean and standard deviation. For sampling-based methods, the batch size was set to 1024, with each hop sampling an equal number of nodes. Regarding hyperedge augmentation, we chose to augment hyperedges by adding a number of nodes equivalent to their original degrees. The implementations were carried out using the PyTorch Geometric library (PyG) \cite{PyG2019} on an NVIDIA A100 40GB GPU.

\section{Results}

Through extensive experimentation, we demonstrate the effectiveness of Adaptive-HGNN by answering the following questions:

\begin{enumerate}
    \item How does our proposed Adaptive-HGNN perform compared with the state-of-the-art hypergraph learning models?

    \item How efficient is our proposed Adaptive-HGNN on memory and training time?

    \item How do the proposed Random Hyperedge Augmentation and Peer-MLP in our Adaptive-HGNN contribute to the overall performance?

    \item Can our proposed Adaptive-HGNN effectively identify and sample informative neighbors for the task?
    
    \item What are the effects of different hyperparameters on the performance of the Adaptive-HGNN framework?

\end{enumerate}


\subsection{Performance Analysis}

\textbf{Performance Comparison}
Table~\ref{tab:hypergraphResult} presents the hypernode classification accuracies for a variety of hypergraph learning models, encompassing both full batch and sampling-based methodologies across a diverse array of datasets. The results distinctly demonstrate the superior performance of our adaptive (Ada) sampling technique in the hypergraph context. The only exception is observed in the DBLP dataset, where the average size of hyperedges, $\overline{|e|} = 4.5$ and the median size, $\widetilde{|e|} = 3$. This indicates that the high-order information is relatively sparse (close to normal graph where $|e|$=2), which may hinder our model's ability to capture it effectively, resulting in comparatively lower performance against state-of-the-art models.
About two large datasets, Trivago and OGBN\_MAG, older baselines like Clique\_{GCN}, Clique\_{GAT}, HNHN, and HGNN are able to manage the computational memory demands of Trivago, but more recent baselines like SetTransformer, ED-HNN and PhenomNN can not scale to it due to their methodology complexity. Notably, all conventional methods falter in scaling to the expansive OGBN\_MAG dataset, which encompasses over 700,000 nodes and 1 million hyperedges. Furthermore, we compared our adaptive sampling methodology against random sampling techniques employing GCN and Graph Transformer architectures. This comparison reveals a marked performance improvement attributable to our adaptive sampling strategy and demonstrates its versatility. Additionally, integrating random Hyperedge Augmentation into our framework has proven to be a beneficial strategy, contributing to a significant uplift in model performance.

\begin{figure}[h]
    \begin{center}
    \includegraphics[width=0.9\linewidth]{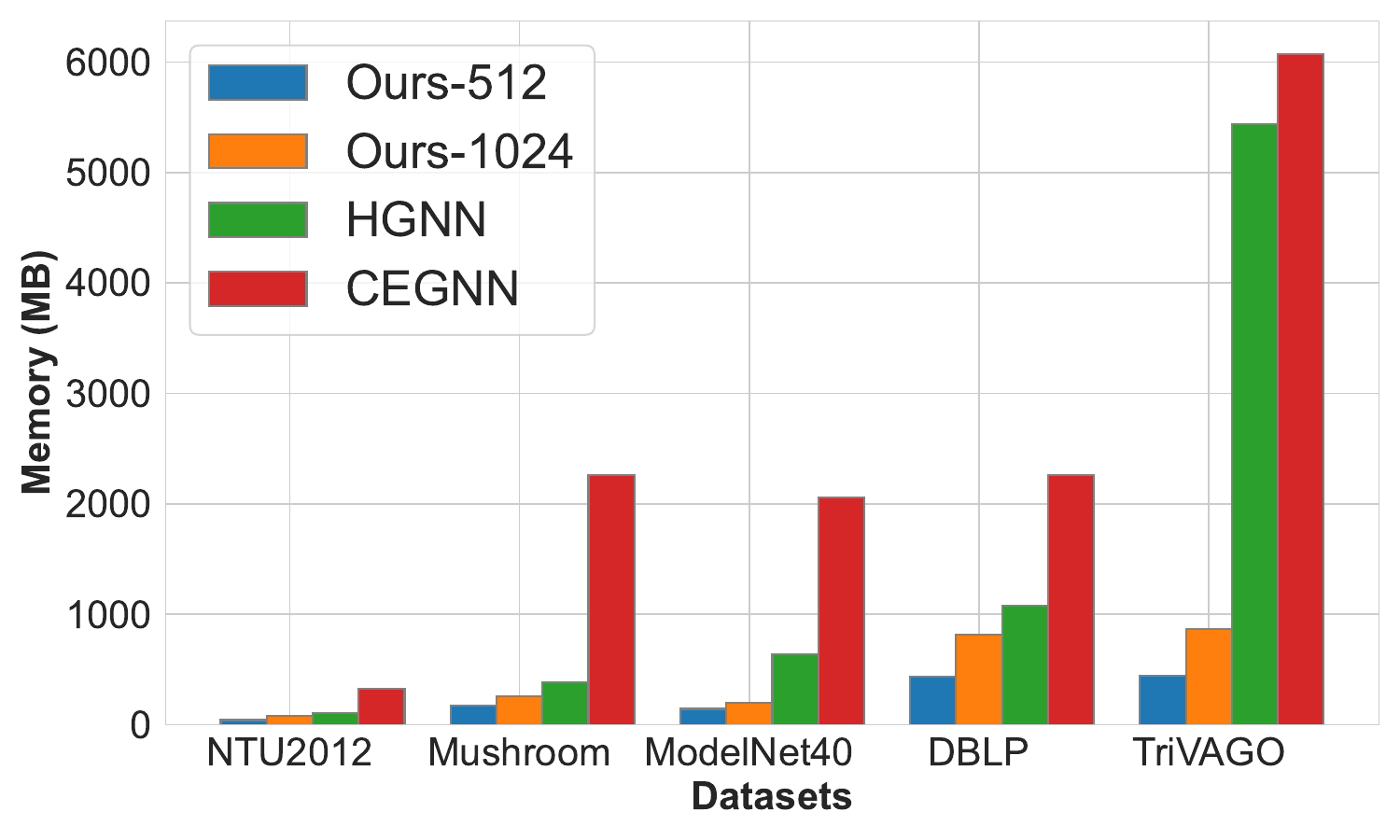}
    \vspace{-3mm}
    \caption{Memory Comparison of our method with 512/1024 batch size, HGNN, and  Clique Expension with GCN(CEGNN).}
    \label{fig:memory}
    \end{center}
    \vspace{-3mm}
\end{figure}

\subsection{Memory and time analysis}
\subsubsection{Memory analysis} As illustrated in Fig.~ \ref{fig:memory}, our proposed methods, utilizing batch sizes of 512 and 1024, significantly reduce memory usage across all datasets compared to HGNN and Clique\_{GCN}, while still achieving superior performance. This reduction in memory consumption is especially pronounced in the context of large-scale datasets like Trivago, where traditional methods, particularly CEGNN, tend to require substantially more memory.

\subsubsection{Time analysis} Table \ref{tab:sampling_comparison} presents a comparison of the accuracy and time between our Adaptive sampling and Random sampling techniques. Our findings demonstrate that the combination of GCN with GFlowNet outperforms random sampling on the ModelNet40 and NTU2012 datasets, resulting in an accuracy improvement of approximately 6\% and 10\%, respectively. This improvement comes at a small increase in training time, with the epoch duration increasing from 0.47 to 0.56, and 0.10 to 0.12 seconds. The performance of GT models with sampling follows a similar trend. With GFlowNet adaptive sampling, we observe higher accuracy, albeit with an increase in training time per epoch.


    

\subsection{Ablation Study}
\subsubsection{Influence of sampling space augmentation}
Table \ref{tab:sampling_comparison} presents a comparative analysis of our method's performance both with and without the implementation of Random Hyperedge Augmentation (RHA). The results highlight the effectiveness of RHA in enhancing our adaptive modeling capabilities. Specifically, for the ModelNet40 dataset, integrating RHA(1)—using a number of nodes equal to the original hyperedge degree—leads to a marked improvement in accuracy, which increases from $93.44\%$ to $97.20\%$. This represents a significant gain of $4.13\%$. Similarly, in the NTU2012 dataset, employing RHA results in an accuracy increase from $82.38\%$ to $84.42\%$, a gain of $3.73\%$. While augmenting with half (0.5) and twice (2) the number of nodes also improves performance, the gains are not as substantial as with a one-time augmentation. These results underscore the critical role of RHA in refining model performance, demonstrating the robustness of our method when enhanced by this sampling space augmentation technique.

\begin{table}[h!] \small
    \centering
    \caption{Ablation study of random (Rdm) and adaptive (Ada) sampling methods on ModelNet and NTU2012 datasets. Graph Convolution Network (GCN)  and Graph Transformer (GT) are used as the backbone. T denotes the training time in seconds per epoch. } 
    \resizebox{\columnwidth}{!}{
    \begin{tabular}{lllll}
    \toprule
    \textbf{Method} & \multicolumn{2}{c}{\textbf{ModelNet40}} & \multicolumn{2}{c}{\textbf{NTU2012}} \\
     & \textbf{Acc} & \textbf{T} & \textbf{Acc} & \textbf{T} \\
    \midrule
    Rdm-GCN             & 86.99 $\pm$ 1.29     & 0.47     & 72.66 $\pm$ 1.40     & 0.10 \\
    Ada-GCN     & 92.53 $\pm$ 0.62     & 0.56     & 83.53 $\pm$ 0.89     & 0.12 \\ \midrule
    Rdm-GT             & 91.28 $\pm$ 1.60     & 0.53     & 74.35 $\pm$ 0.77     & 0.13 \\
    Ada-GT            & 93.44 $\pm$ 0.67     & 0.74     & 82.38 $\pm$ 0.75     & 0.16 \\
    Ada-GT + RHA(0.5)            & 96.42 $\pm$0.25     & 0.79     & 83.14 $\pm$ 1.24    & 0.17 \\
    Ada-GT + RHA(1)            & \textbf{97.20 $\pm$0.39}     & 0.82     & \textbf{84.42 $\pm$ 1.39}    & 0.19 \\
    Ada-GT + RHA(2)           & 96.47 $\pm$0.13    & 0.84     & 84.32 $\pm$ 1.07    & 0.20 \\
     \bottomrule
    \end{tabular}
    }
\label{tab:sampling_comparison}
\end{table}

\subsubsection{Influence of MLP-enhanced intialization}
Table~\ref{tab:MLP_init} demonstrates the impact of MLP initialization on hypergraph learning. In this experiment, we transferred weights directly from a peer MLP (only trained on node features) to a GCN model and assessed the inference performance against GCN models that were randomly initialized and untrained. We observe that even without further training, MLP\_init-GCN achieves a performance of $90.37$ and $81.65$, which is close to the final performance of $92.53$ and $83.53$ on ModelNet40 and NTU2012, respectively. In practice, MLPs require fewer than 50 epochs to converge, rendering the MLP training duration in the MLPInit method almost negligible compared to that of GNNs. This not only streamlines the entire training process but also offers a substantial reduction in time and resource expenditure compared with training from scratch.

\begin{table}[h]
\setlength\tabcolsep{13pt}
\centering
\caption{The inference performance of GCN with random initialized weights and weights from trained MLP}
    \begin{tabular}{lll}
         \toprule
        Method             & ModelNet40         & NTU2012           \\ \hline
        Rdm\_init-GCN         &  2.65 $\pm$ 2.46    &  1.55 $\pm$ 1.50   \\
        MLP\_init-GCN         & 90.37 $\pm$ 0.42   & 81.65 $\pm$ 1.28  \\
        Gain               & 87.72              & 80.10              \\ \hline
        Final performance  & 92.53 $\pm$ 0.62   & 83.53 $\pm$ 0.89  \\  \bottomrule
    \end{tabular}
\label{tab:MLP_init}
\end{table}

\subsection{Informative neighbors sampling}
\subsubsection{Learning sample preference} 
\begin{figure}[h]
    \begin{center}
    \includegraphics[width=\linewidth]{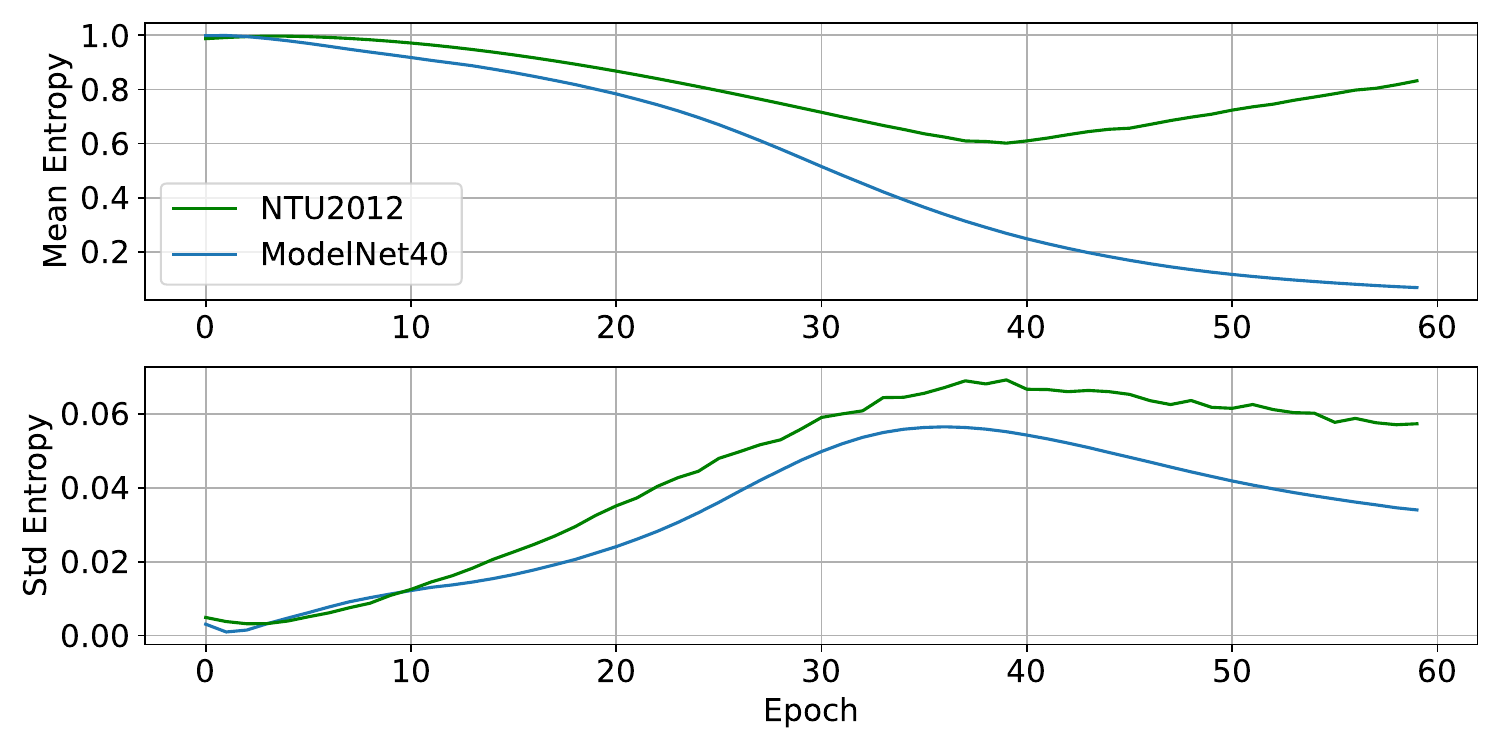}
    \vspace{-6mm}
    \caption{Mean and standard deviation entropy of nodes including probability output by adaptive sampling network. Small means represents Ada-GT learns strong preference to include/exclude some nodes}
    \vspace{-2mm}
    \label{fig:entropy}
    \end{center}
\end{figure}

To see if our approach, Ada-GT, can learn the preference pattern over node sampling, we calculate the entropy of sampling following ~\cite{younesian2023grapes}. Figure~\ref{fig:entropy} depicts the progression of both the mean and standard deviation of the node's sampling preference entropy output by policy network as training progresses across epochs. Entropy, in this context, measures the uncertainty in the probability distribution output by the GFlowNet model for node sampling. The top graph shows the mean entropy of the nodes, which initially decreases and then stabilizes over epochs for both datasets. Notably, the NTU2012 dataset shows a higher mean entropy compared to ModelNet40. This stability indicates that the GFlowNet model consistently maintains a certain level of uncertainty in its sampling decisions throughout training, which can be beneficial for adequately exploring the diversity of solution space. The bottom figure plots the standard deviation of the entropy, reflecting the variability in the entropy of individual nodes. For both datasets, we observe an increase in the standard deviation as training progresses, followed by convergence. The small standard deviation at the beginning of training suggests a relatively uniform sampling across nodes, whereas the increasing trend points towards the development of a strong preference to include or not include certain nodes as training progresses. This behavior implies that the GFlowNet model adapts its sampling strategy to focus more on certain groups of nodes of the hypergraph, potentially honing in on regions that contribute more to learning the task at hand.
\begin{table*}[ht]
\setlength\tabcolsep{12pt}
\centering
\caption{Experimenting with Various Training Split Sizes for Hypernode Classification. OOM indicates out-of-memory.}
\label{tab:experiment_results}
\begin{tabular}{lllllllll}
\hline
\textbf{Train \%} & \textbf{Model} & \textbf{20News} & \textbf{Mushroom} & \textbf{ModelNet40} & \textbf{NTU2012} & \textbf{TriVago} \\ \hline
\multirow{4}{*}{10\%} & $\mbox{Clique}_{GCN}$ & OOM & 91.02 ± 0.34 & 85.30 ± 0.49 & 64.74 ± 2.09  & 14.79 ± 2.79 \\
                      & $\mbox{Clique}_{GAT}$  & OOM & 82.64 ± 0.42 & 82.38 ± 0.88 & 58.48 ± 2.01  & 15.72 ± 1.79 \\
                      & HGNN & 71.16 ± 0.44 & 92.84 ± 0.50 & OOM & 62.88 ± 0.71  & 14.51 ± 3.34 \\
                      & \textit{ours} & \textbf{77.63 ± 0.59} & \textbf{99.66 ± 0.22} & \textbf{95.60 ± 0.31} & \textbf{66.32 ± 1.67}  & \textbf{27.42 ± 0.42} \\ \hline

\multirow{4}{*}{20\%} & $\mbox{Clique}_{GCN}$ & OOM & 91.03 ± 0.33 & 85.63 ± 0.29 & 69.77 ± 2.08 & 14.90 ± 2.34 \\
                      & $\mbox{Clique}_{GAT}$  & OOM & 82.82 ± 0.53 & 83.03 ± 1.15 & 64.64 ± 1.35 & 16.36 ± 2.47 \\
                      & HGNN & 70.69 ± 0.55 & 93.15 ± 0.34 & 88.95 ± 0.60 & 70.75 ± 2.32 & 11.41 ± 0.98 \\
                      & \textit{ours} & \textbf{78.78 ± 0.55} & \textbf{99.93 ± 0.13} & \textbf{96.37 ± 0.17} & \textbf{77.40 ± 0.94} & \textbf{31.82 ± 0.23} \\ \hline

\multirow{4}{*}{40\%} & $\mbox{Clique}_{GCN}$     & OOM             & 91.25 $\pm$ 0.12       & 85.58 $\pm$ 0.12     &  71.72 $\pm$ 4.44        & 13.92 $\pm$ 3.37 \\ 
                      & $\mbox{Clique}_{GAT}$     & OOM             &  83.51 $\pm$ 0.85      &  83.51 $\pm$ 0.85    &  63.76 $\pm$ 2.11         & 14.82$\pm$ 1.92  \\
                      & HGNN                      & 78.31 $\pm$ 0.55   &93.59 $\pm$ 0.40      &90.99 $\pm$ 0.03     &  73.57 $\pm$ 0.50         &26.58 $\pm$ 0.34 \\ 
                      & \textit{ours}    & $\mathbf{79.78 \pm 0.72}$     &  $\mathbf{100 \pm 0.00}$   & $\mathbf{97.20 \pm 0.39}$ &   $\mathbf{84.42 \pm 1.39}$           &  $\mathbf{39.42 \pm 0.63}$ \\
\hline
\label{tab:Hyperparameter}

\end{tabular}

\end{table*}

\subsubsection{Case Study} 
\begin{figure}[h]
    \begin{center}
    \vspace{-5mm}
    \includegraphics[trim={6cm 9cm 10cm 6cm},clip,width=0.8\linewidth]{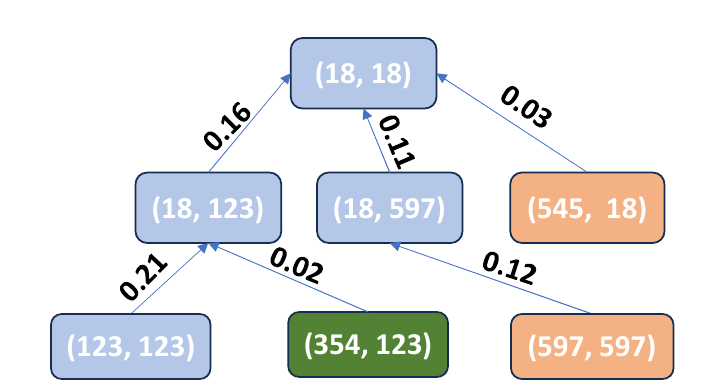}
    \vspace{-2mm}
    \caption{Node (18,18) and its subset of neighbors from the ModelNet40 dataset. The Adaptive-HGNN model identifies which neighbors are informative. The numbers in the nodes represent the hypernode\_id and hyperedge\_id. Colors indicate the class of each node}
    \vspace{-2mm}
    \label{fig:case}
    \end{center}
\end{figure}

In Fig. \ref{fig:case}, we find the Ada-HGNN effectively discriminates between informative and less informative neighbors. For the task of node classification, the effective model normally assigns high sampling probabilities to neighbors with the same labels and lower probabilities to those with different labels~\cite{zhu2020beyond}. For instance, for the target node (18,18), the policy network outputs a higher preference for node (18,123) with a sampling probability of 0.16, whereas it assigns a significantly lower probability of 0.03 to node (545, 18). Additionally, in a two-hop sampling scenario, node (123,123) is shown to have a high probability of passing its information to node (18,18), illustrating the adaptive capability of our model to prioritize and process information based on node class for node classification dataset.

\begin{figure}[h]
    \begin{center}
    \includegraphics[width=0.9\linewidth]{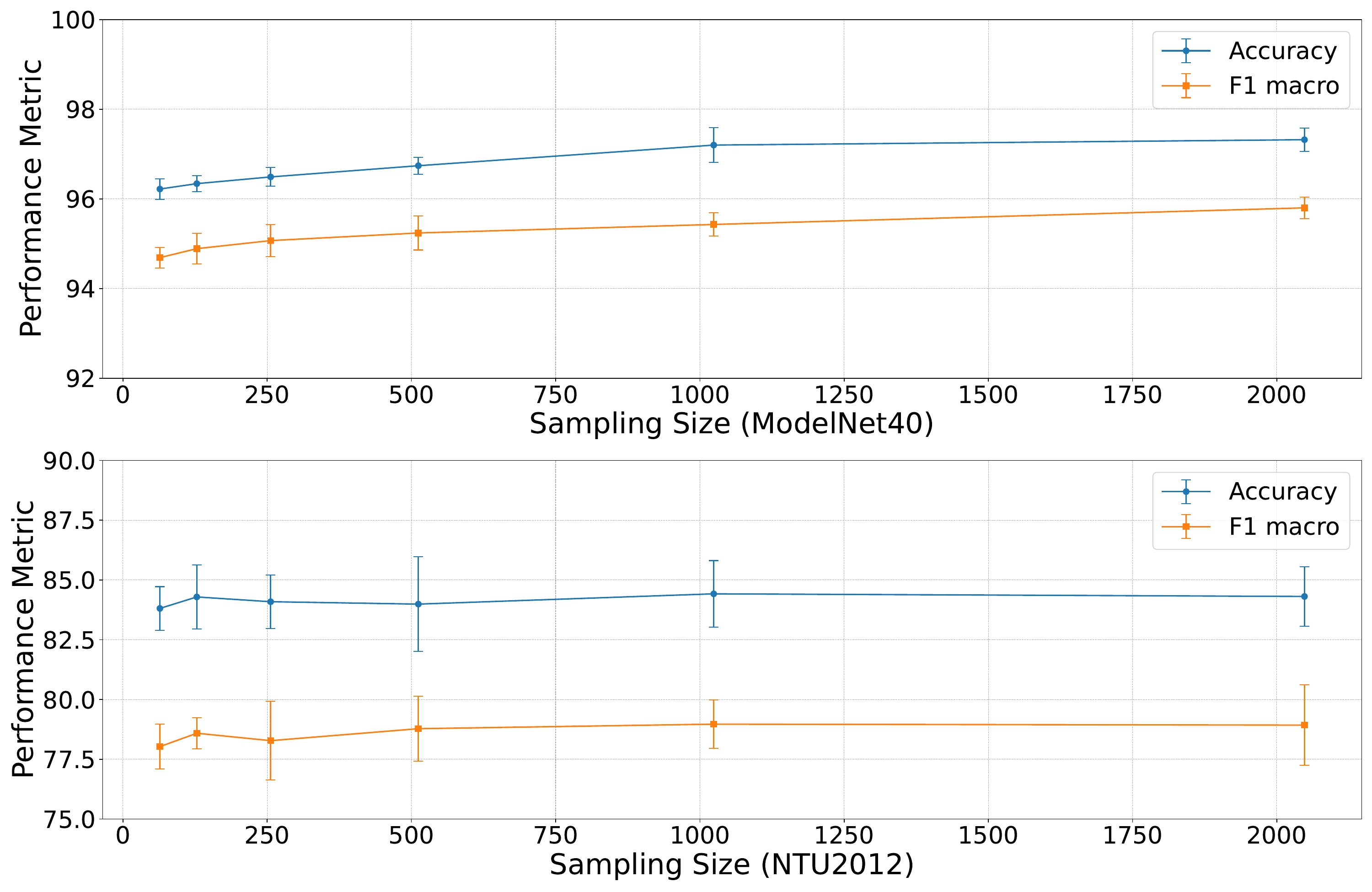}
    \vspace{-3mm}
    \caption{  The Impact of Varying Sampling Sizes on Hypernode Classification Performance for the ModelNet40 and NTU2012 Datasets. We assess performance using accuracy and F1 score metrics, with error bars indicating the standard deviation across five trials}
    \vspace{-2mm}
    \label{fig:Sensitivity}
    \end{center}
\end{figure}

\subsection{Hyperparameter Sensitity} 

\subsubsection{Training split size} We conducted an experiment where we trained our model with significantly reduced sizes of the training set. Specifically, we explored the performance of our model under stringent constraints where only $10\%$, $20\%$, and $40\%$. The results of this experiment, as shown in the Table ~\ref{tab:Hyperparameter}, suggest that our model maintains robust performance even with a reduced training set size. For instance, at the $10\%$ training data size, our model achieved an accuracy of $77.63 \pm 0.59$ on the 20News dataset and $99.66\pm0.22$ on the Mushroom dataset. At the $20\%$ training data size, there is a slight improvement in performance, which is expected due to the larger training size. However, the key observation is that even with only $10\%$ of the data for training, our model's performance is competitive, indicating its effectiveness in leveraging small samples for training. This ability to learn effectively from a small sample size is particularly important for scenarios where data collection is expensive or where the available data is limited. The experiment confirms that our proposed method has practical relevance and could be a valuable tool in such contexts.

 \subsubsection{Sampling node number} As shown in Fig.~\ref{fig:Sensitivity} we have systematically investigated the influence of different node sampling sizes on our model's accuracy. The experimental results demonstrate not only the method’s performance across various sampling sizes but also its robustness. On ModelNet40, we observe a small increase in accuracy with larger sampling sizes, peaking at a sampling size of 2048 and performing well on a very low sampling size of 64.  Similar patterns of robustness are evident in the NTU2012 dataset, with the model exhibiting stable performance despite the variation in sampling size. These findings affirm the method’s resilience to hyperparameter variations, reinforcing its reliability and practicality for different dataset complexities.

%% file: parts/05-conclusion.tex


\section{Conclusion}
In conclusion, this study presents a comprehensive approach to addressing scalability challenges in HGNNs and enhancing their effectiveness for large-scale applications. By introducing a novel adaptive sampling technique specifically designed for hypergraphs, we effectively capture the complex relationships between nodes and hyperedges. Our method integrates reward-driven learning to retain essential hypergraph features while mitigating overfitting and ensuring diversity. Additionally, the incorporation of random hyperedge augmentation and a supplementary MLP module enhances robustness and generalization. Extensive experiments on real-world datasets demonstrate the superiority of our approach, significantly reducing computational and memory costs while outperforming traditional HGNNs and baseline methods in node classification tasks. Overall, this work contributes to the advancement of hypergraph learning architectures, offering a promising pathway for enhancing the scalability and effectiveness of HGNNs in diverse applications.